# Simulated Tom Thumb, the Rule Of Thumb for Autonomous Robots

**M. A. El-Dosuky [1], M. Z. Rashad [1], T. T. Hamza [1], and A.H. EL-Bassiouny [2]**

[1] Department of Computer Sciences, Faculty of Computers and Information sciences, Mansoura University, Egypt

[2] Department of Mathematics, Faculty of Sciences, Mansoura University, Egypt

## Abstract

For a mobile robot to be truly autonomous, it must solve the simultaneous localization and mapping (SLAM) problem. We develop a new metaheuristic algorithm called Simulated Tom Thumb (STT), based on the detailed adventure of the clever Tom Thumb and advances in researches relating to path planning based on potential functions. Investigations show that it is very promising and could be seen as an optimization of the powerful solution of SLAM with data association and learning capabilities. STT outperform JCBB. The performance is 100 % match.

## 1. Introduction

For a mobile robot to be autonomous, it must solve the simultaneous localization and mapping (SLAM) problem, i.e. to be able to progressively build a consistent map of an unknown environment while simultaneously determining its location within this dynamically changing map [6]. SLAM necessitates updating a complex state of the pose and all observed landmark positions [10]. Many models are proposed solving SLAM, mainly the extended Kalman filter (EKF-SLAM) and Rao-Blackwellized particle filters (FastSLAM) both are best explained in [6]. Implementations exploit visual [1] or sonar navigation [4]. Data association is very crucial task to find matches between the current landmarks and the new set of observed features [12]. In stochastic mapping [20], data association is simply addressed using a classical technique in tracking problems called the nearest neighbor (NN) [14]. The development of Joint Compatibility Branch-and-Bound data association (JCBB) [12] is encouraged by the fact that the simple Nearest Neighbor (NN) algorithm for data association is very sensitive to the sensor error, increasing the probability of matching unrelated map features [12]. JCBB employs a validation mechanism that determines the joint compatibility of a set of pairings in the hypothesis. This algorithm is robust than NN algorithms and feasible in terms of computational cost [12].

The general architecture of control unit of most autonomous robots is shown in the next figure [18].

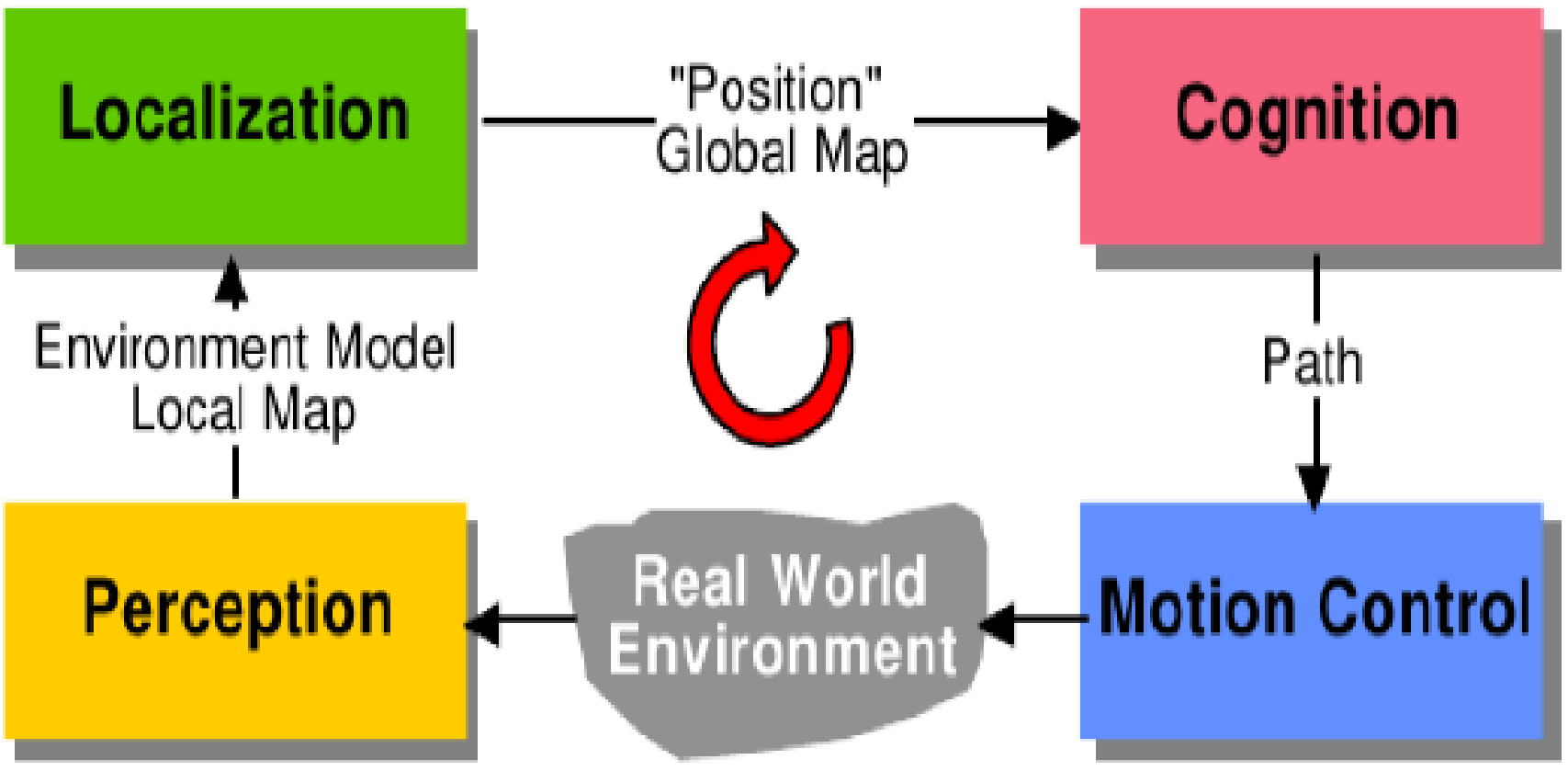

Figure 1 Control Architecture

Initially, a robot can run a random search such as Lévy flights [12]. Then, for navigation and path planning, a robot can use artificial potential functions as spike-timing dependent plasticity (STDP) [15]. For environment representation, there are three forms: Continuous Metric, Discrete Metric (also known as Metric grid) and Discrete Topological (or Topological grid)[18] .

We develop a new metaheuristic algorithm called Simulated Tom Thumb (STT), based on the detailed adventure of the clever Tom Thumb and advances in researches relating to path planning based on potential functions. Section 2 reviews the interesting nursery story of Le petit Poucet, the French version of Tom Thumb and proposes the algorithm. Section 3 evaluates the algorithm, comparing it with the implementation of [11]. Investigations show that it is very promising and could be seen as an optimization of the powerful solution of SLAM with data association and learning capabilities.

## 2. Simulated Tom Thumb

Let us first review the interesting nursery story of Le petit Poucet, the French version of Tom Thumb, to facilitate describing the proposed algorithm.

### 2.1 The Adventure of Tom Thumb

Once upon a time, Tom Thumb was a child born to a very poor family with seven children. Tom Thumb heard his parents deciding to abandon them in the forest. He had the chance to collect little stones and dropped them as a trail on their way. After the parents fled, he could successfully lead the children back home. The next time the parents decided to abandon them; they prevented him from collecting stones, so he took some bread. Tom Thumb scattered crumbs as a trail, but these had disappeared. After the parents fled, children followed the light of the ogre's house. The kind wife of the ogre welcomed them. She took them to her daughters' room where there were two beds: one for the girls, the other for the boys. Tom Thumb exchanged the crowns on girls' heads with the hats of the boys to trick the ogre … Tom Thumb took the magical seven-mile boots of the ogre that allowed him to jump over mountains and to be hired by the king as a messenger. He was rewarded a lot of money and returned home with his family who stopped abandoning him!

Drogoul and Freber simulate the first part of the story into three types of Tom Thumb robot [5]. Tzafestas argues that Tom Thumb robot as defined in [5] is not stable because it assumes unbounded number of crumbs, which is not physically possible in real implementation [19]. Tzafestas proposed a laydown-pickup mechanism that a trail to a source is built quickly and reinforced as long the source exists and vanishes shortly after the the source is exhausted. This regulation mechanism ensures that no agent will ever roun out of crumbs. This regulation has two problems [19]. First, it depends on a global view of the environment to identify the prober paramter for each situation. With approximating equations, it loses its global optimization characteristic. Second, linear regulation is not a learning model.

### 2.2 Lévy flight foraging

Predators' foraging path is a random walk because the next move decision depends on the current location and the transition probability to the next location [13]. Many recent studies assure two patterns of random walks. They found that Brownian movement is associated with abundant preys and Lévy flight is associated with sparser or unpredictably distributed preys [2]. With apparent potential capability, Lévy flight is applied to optimal search [13].

Lévy flight is applied to generate new states $s^{(t+1)}$ based on current state.

$$s^{(t+1)} = s^{(t)} + \alpha * \text{Lévy}\,(\lambda) \qquad (1)$$

where $\alpha > 0$ is the step size, we can use $\alpha = O(1)$. The product $*$ means entry-wise multiplications. Lévy flights obey a Lévy distribution for large steps

$$\text{Lévy} \sim u = t^{-\lambda}, \quad (1 < \lambda \leq 3) \quad (2)$$

which has an infinite variance with an infinite mean. But for simplicity, jumps of a robot can be made to obey a step-length distribution with a heavy tail.

**2. 3 Artificial Sensorimotor Learning**

Hebbian learning, an algorithm for altering connection weights in neural networks, is only sensitive to the spatial correlations between pre- and postsynaptic neurons [*3*]. Spike-timing dependent plasticity (STDP) is temporally asymmetric form of Hebbian learning [9]. According to the STDP rule, synaptic weight $w_{ij}$ between a pre-synaptic neuron *j* and a neuron *i* after a time step $\Delta t$ is, [*7*]:

$$w_{ij}(t+\Delta t) = w_{ij}(t) + \Delta w_{ij}(t) \quad (3)$$

Mobile robots can adapt movements based on sensor data. Let $W \subset \Re^3$ denote an Euclidean robot workspace with a set of fixed obstacles $\{B_j\}_{j \in I_B}$. First, with no prior knowledge of obstacles SLAM is required for the initial partially-observable environment. Next, robot must take into account feedback controllers to deal with partially-unmodeled dynamics.

We adopt the geometry approach developed by Ferrari in [8] and SLAM to model FOV, $S \subset \Re^3$. Figure 2 shows an example of these geometries [8]

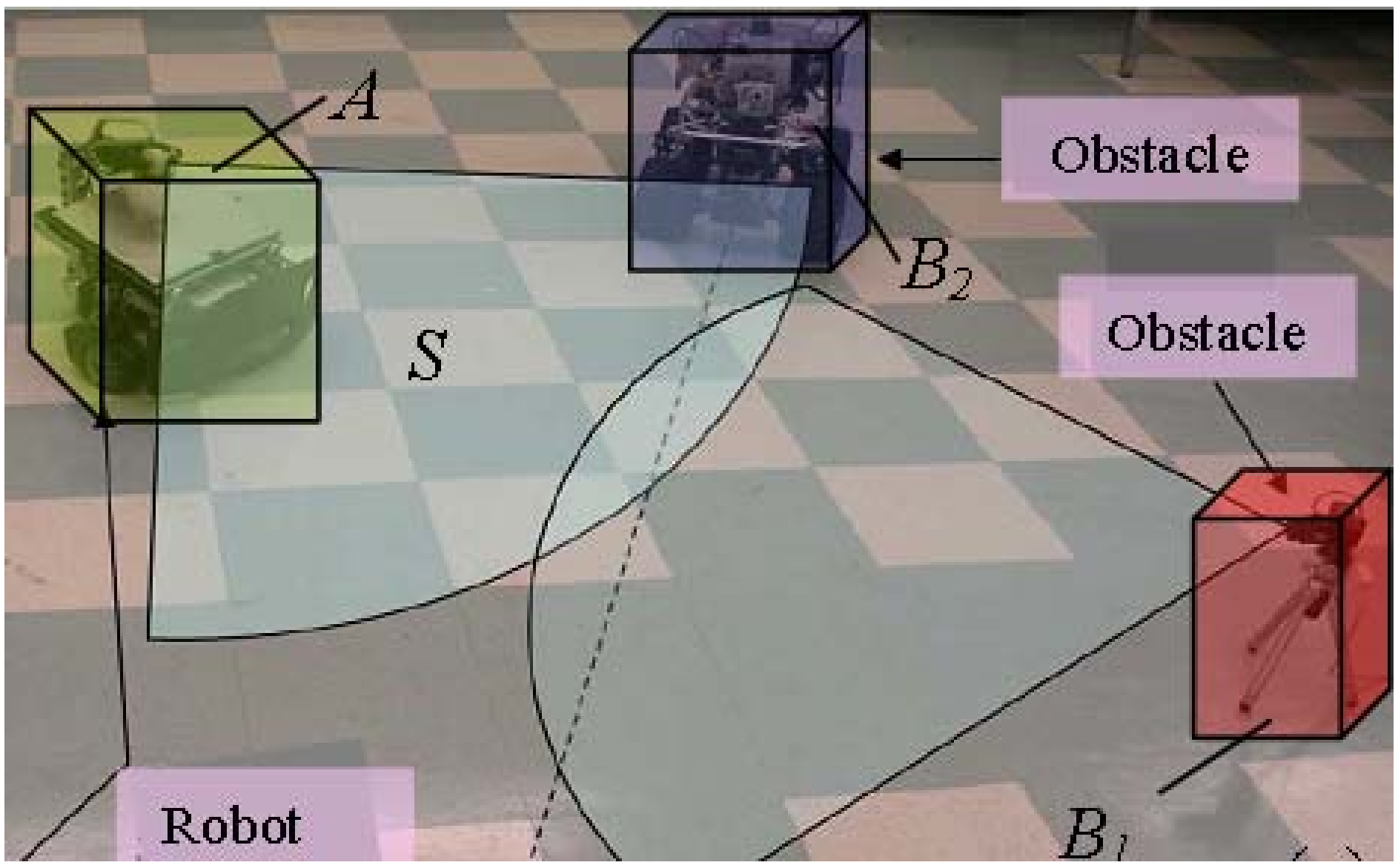

Figure 2 Geometric models of robot, *A*, sensor FOV, *S*, and two obstacles, $B_1$ and $B_2$

A robot is a dynamical system that obeys a nonlinear ODE of the form,

$$\dot{x}(t) = f[x(t), u(t), t], \text{ with } x \in X \subset \Re^{n\times 1}, \text{ and } u \in U \subset \Re^{m\times 1}. \quad (4)$$

Where, *x* is the state, *u* is control inputs, $(\dot{\cdot})$ denotes the derivative with respect to time. The macroscopic behavioral goals can be expressed by the value function or *cost-to-go*,

$$V^{\pi}[x(t_k)] = \sum_{t_k = t}^{t_f - 1} \mathscr{L}[x(t_k), u(t_k), x(t_k + 1)], \text{ with } u(t_k) = \pi[x(t_k)]. \quad (5)$$

which is the future performance within $[t_k, t_f]$ time interval, subject to the present control law $\pi(\cdot)$. The Lagrangian $\mathscr{L}[\cdot]$ represents instantaneous behavioral goals as a function of *x* and *u*.

### 2.4 Proposed Simulated Tom Thumb

Let us summarize up the adventure of Tom Thumb to the following general assumptions:

A1. The family consists of the parents, Tom Thumb and seven children.
A2. Initially, family is very poor.
A3. As long as the family is poor, parents will abandon the children.
A4. Children obey their parents; even they ask them to follow them to the forest.
A5. Parents go to the forest using random walks, and children learn.
A6. Parents can allow Tom Thumb to collect little stones.
A7. If parents prevented Tom Thumb from collecting stones, he took some bread.
A8. Tom Thumb drops stones/crumbs as a trail on their way to the forest.
A9. If parents flee, children obey Tom Thumb to lead them back home.
A10. Tom Thumb follows the trail until it is disappeared.
A11. If the trail is disappeared, children will apply a random walk, learned from A5.
A12. All children are boys. All the ogre's children are girls.
A13. All boys wear hats, All girls' wear crowns
A14. At the ogre's house, Tom Thumb exchanges hats with crowns.
A15. Tom Thumb takes the magical seven-mile boots of the ogre
A16. Magical boots allows to jump over mountains, i.e. faster random walks
A17. Hired by the king, Tom Thumb is rewarded a lot of money and returns home.

Minor negotiable detailed assumptions, for varying implantations, are:

B1. They explore the landscape randomly using Lévy flight path;
B2. Children learn using the STDP rule
B3. Heads' wear differentiates boys from girls. This may be simulated by different sign
B4. Family's field-of-view (FOV) is wide and is divided into 3x3 grid cells as seen in figure 3.

| | | |
|---|---|---|
| parents | $brother_1$ | $brother_2$ |
| $brother_3$ | Tom | $brother_5$ |
| $brother_4$ | $brother_7$ | $brother_6$ |

Figure 3. Family's field-of-view, divided into 3x3 grid cells

Based on these assumptions, the Simulated Tom Thumb algorithm is shown in figure 4.

FUNCTION **Simulated_Tom_Thumb**()
Step1. Create the family WINDOW of the parents, Tom Thumb and seven children.
Step2. Generate random environment GRID
Step3. Populate mountains, king palace, and ogre's house randomly
Step4. Set current location to HOME
Step5. Set family's wallet to be EMPTY, $\alpha$ the step size to 1.
Step6. WHILE family's wallet is EMPTY DO steps 7 – 13
- Step7. Parents allow or prevent Tom Thumb to collect little stones
- Step8. If Tom Thumb is prevented from collecting stones, he took some bread
- Step9. Set heads' wear to HAT=1
- Step10. Move family WINDOW go to the forest using Lévy flight
- Step11. Tom Thumb drops stones/crumbs as a trail on their way to the forest
- Step12. Children learn seen features and walk way using STDP rule.

Step13. If parents flee, do steps 14 – 16
Step14. Zero the parent cell from the family WINDOW
Step15. Tom Thumb follows the trail until it is disappeared.
Step16. If the trail is disappeared, do steps 17 – 19
Step17. Random walk, sensing around the family WINDOW
Step18. If current location is HOME, end this iteration
Step19. If current location is ogre's house, do steps 20 – 24
Step20. Set heads' wear to CROWN= - 1
Step21. Magical boots maximize $\alpha$, the step size
Step22. Faster random walks sensing around the family WINDOW
Step23. If current location is HOME, end this iteration
Step24. If current location is the king palace, do steps 25 – 27
Step25. Let the king move the family WINDOW to HOME.
Step26. Let the king set the family's wallet with a numeric award
Step27. If the award is INFENITY, stop

Figure 4 Simulated Tom Thumb

Mountains mentioned in step 3, can be a metaphor for local maxima, to ensure a non-smooth landscape. King palace is a region with special mark, to denote a desired target. Ogre's house is a region with negative sign. In step 7, it is preferable that parents allow Tom Thumb to collect little stones, at least for the first iteration to force a learning session. The difference between stones and crumbs is that crumbs are easy to be forgotten. This is achieved by modifying the change in weight $\Delta w_{ij}$ . Magical boots allows performing faster random walks or jumping over mountains. Many numerical methods are developed to find the greatest or least value of a function [16].

## 3 Evaluations

### 3.1 Experimental Results

José Neira builds an EKF-SLAM simulation that demonstrates JCBB data association in MATLAB™ [11]. A mobile robot carries out a square trajectory in an environment with point features at each side of the trajectory, similar to a cloister as seen in figure 5 where red points and trajectory are ground truth. The vehicle is equipped with a point detector whose characteristics (range, precision) can be modified.

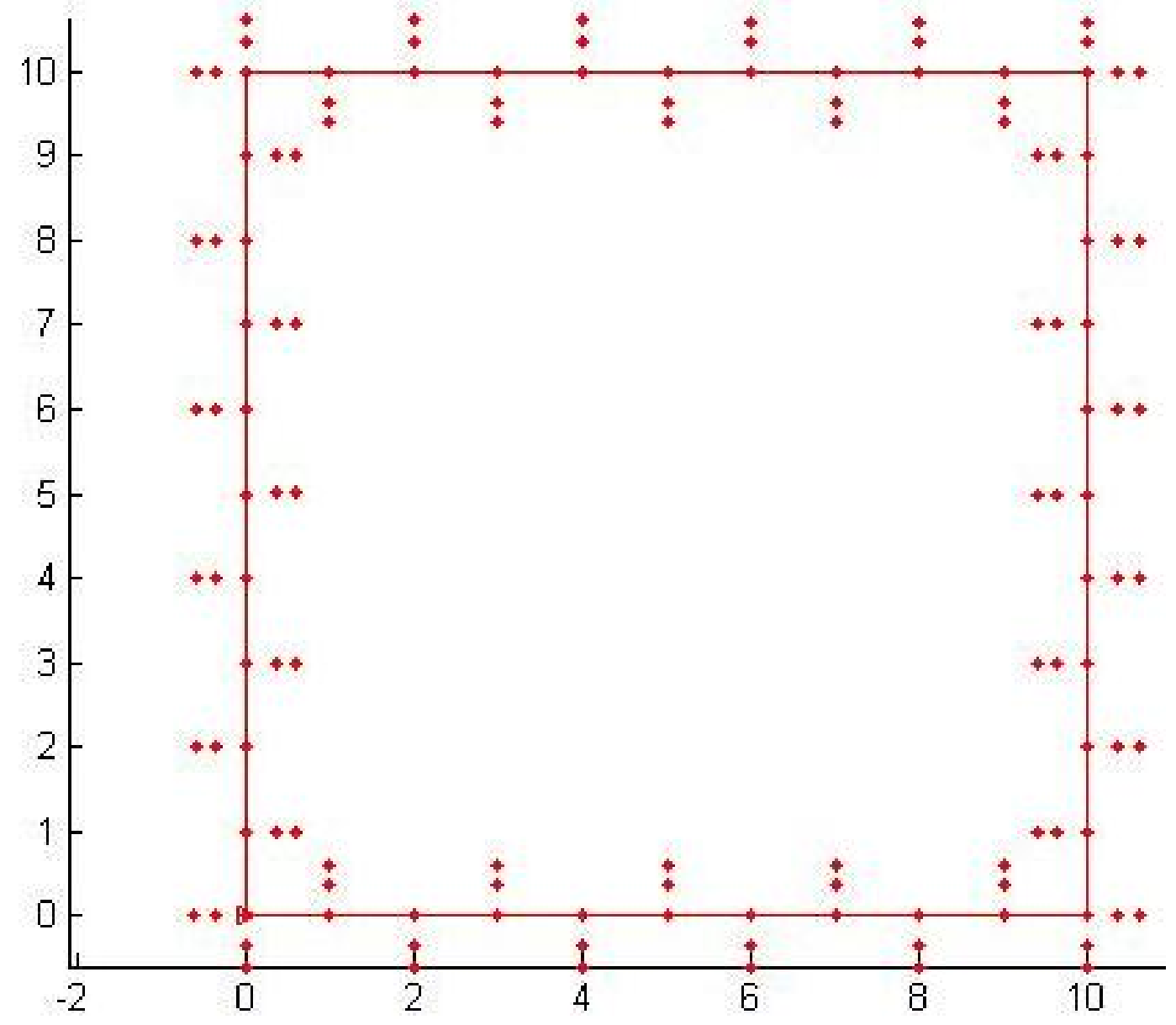

**Figure 5** ground truth

Figure 6 shows the trail and the depiction of error in x and y directions by exploiting JCBB data association. We modified the simulation in [11] to build STT function and the corresponding trail and the depiction of error in x and y are shown in figure 7.

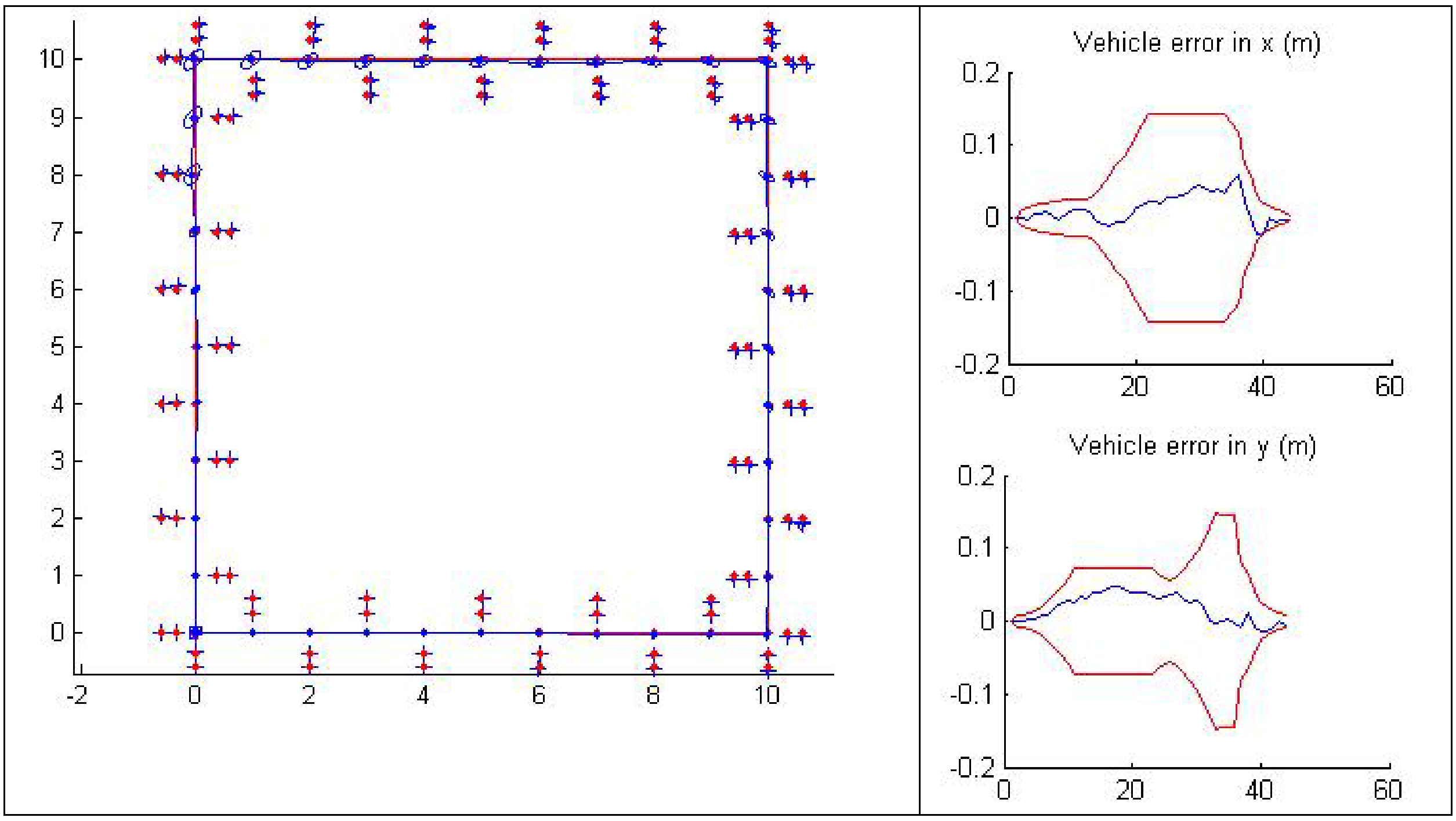

Figure 6. Trail and the error using JCBB data association

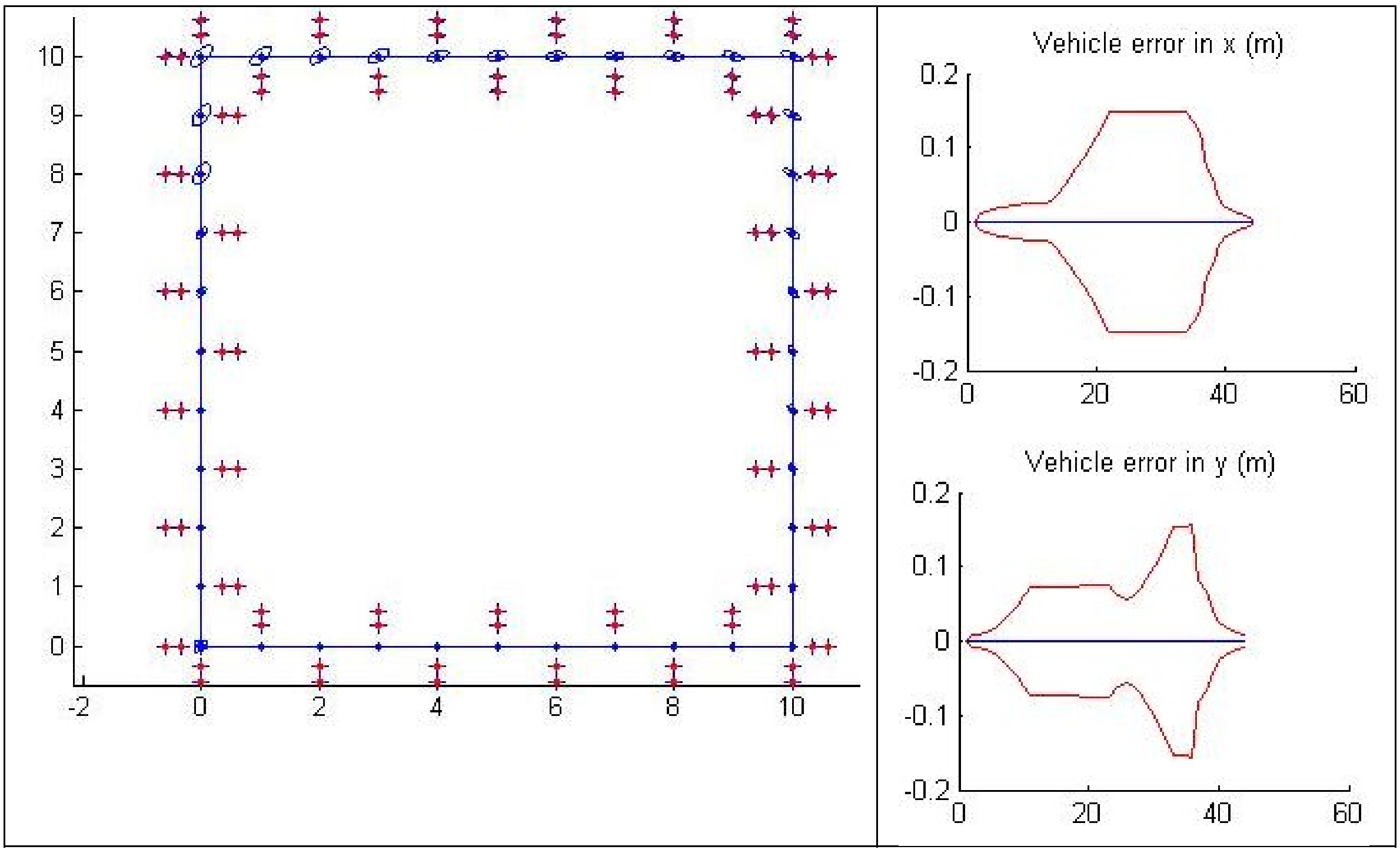

Figure 7. Trail and the error using STT

### 3.2. Discussions and Conclusions

By comparing figure 6 and figure 7, it is clear that STT outperform JCBB. We force the first iteration to make the parent teach their children how to follow the ground truth. The performance is 100 % match. Even without this teaching session, the sliding family WINDOW is enough to filter noise with matching rate over 96%. We believe that a robot designer shall give his robot a lesson or two in vitro before going in vivo. This assumption guarantees that the mobile robot will apply what it learns.